\documentclass[sigconf]{acmart}
\pdfoutput=1

\usepackage{booktabs} 
\setcopyright{acmcopyright}

\acmDOI{}
\acmISBN{}
\acmConference[KDD Workshop on AI for fashion]{KDD}{August 2018}{London, United Kingdom} 
\acmYear{2018}
\copyrightyear{2018}
\acmPrice{}

\begin{document}
\title[Outfit Generation and Style Extraction]{Outfit Generation and Style Extraction via Bidirectional LSTM and Autoencoder}

\author{Takuma Nakamura}
\affiliation{
  \institution{ZOZO Research}
  \city{Tokyo}
  \state{Japan}
}
\email{takuma.nakamura@zozo.com}

\author{Ryosuke Goto}
\affiliation{
  \institution{ZOZO Research}
  \city{Tokyo}
  \state{Japan}
}
\email{ryosuke.goto@zozo.com}

\begin{abstract}
When creating an outfit, style is a criterion in selecting each fashion item. This means that style can be regarded as a feature of the overall outfit. However, in various previous studies on outfit generation, there have been few methods focusing on global information obtained from an outfit. To address this deficiency, we have incorporated an unsupervised style extraction module into a model to learn outfits. Using the style information of an outfit as a whole, the proposed model succeeded in generating outfits more flexibly without requiring additional information
. Moreover, the style information extracted by the proposed model is easy to interpret. The proposed model was evaluated on two human-generated outfit datasets. In a fashion item prediction task (missing prediction task), the proposed model outperformed a baseline method. 
In a style extraction task, the proposed model extracted some easily distinguishable styles.
In an outfit generation task, the proposed model generated an outfit while controlling its styles
. This capability allows us to generate fashionable outfits according to various preferences.
\end{abstract}



\maketitle
\section{INTRODUCTION}
Choosing everyday clothes smartly and purchasing new fashion items are difficult for many people because they require expert knowledge about the characteristics and combinations of fashion items.
Because fashion items are very diverse with variable values, most people need the help of some systems.
To develop systems that assist such tasks, we should solve two important problems: fashion compatibility and fashion style recognition.

Learning compatibility between fashion items is an essential problem in the fashion domain. Fashion compatibility is a criterion for judging whether a pair or group of items are suitable or not. Many researchers have studied fashion compatibility. In their studies, one of the mainstream methods is metric learning, which is achieved by selecting co-purchased items and an outfit dataset and calculating distances between selected items~\cite{vasileva2018learning,shankar2017deep,veit2015learning}. In some studies, compatibility has been learnt by regarding items in an outfit as a sequence~\cite{han2017learning}. The problem with these studies lies in the inability to apply fashion styles that determine the taste of an outfit because these studies only focus on local pairwise relationships.

Another challenge is to capture the underlying fashion style when considering outfits as combinations of fashion items. Fashion styles, such as rock and mode, represent different principles of organizing outfits. 

To realize an advanced fashion recommender system, we should focus not only on estimating personal preferences but also on interpreting fashion styles. In the research of understanding fashion style, Takagi et al.~\cite{takagi2017makes} gathered and learnt to snap images that are labeled as a typical fashion style. However, some outfits have an ambiguous style and cannot be classified. In other studies, fashion styles have been extracted with unsupervised learning using snap images with attributes~\cite{hsiao2017learning}. These studies focused on understanding and extracting the fashion style concept. Thus, they cannot be applied to recommendations and outfit generation tasks.

\begin{figure}
\includegraphics[width=\linewidth]{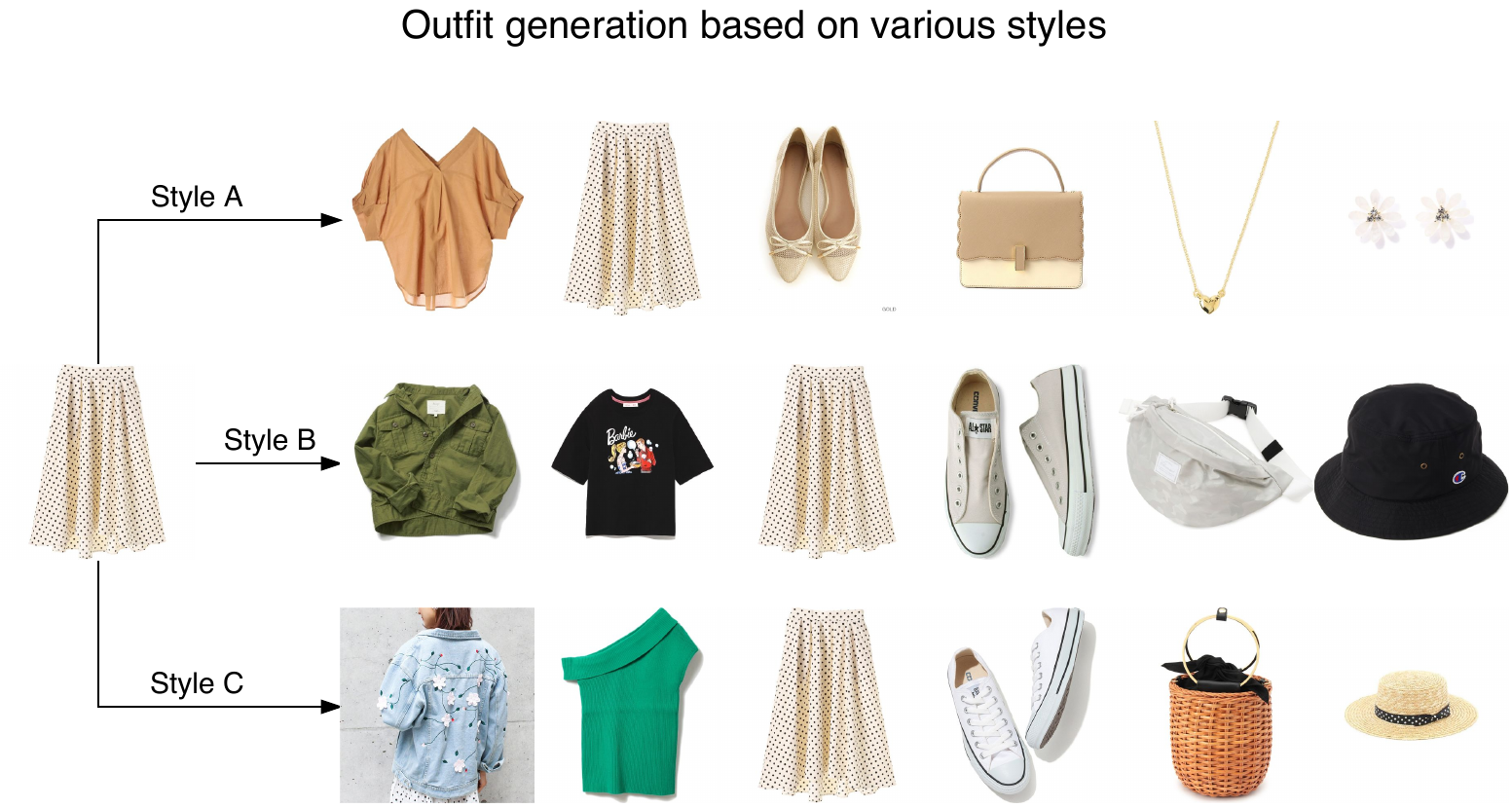}
\caption{Given a query item and some target styles, the proposed model generates different outfits according to the target styles. The proposed model evaluates both item compatibility and the overall style of the items.}
\label{fig:example}
\end{figure}

Motivated by the previous research, we propose a model that learns outfit compatibility and extracts the outfit style simultaneously. These abilities are implemented by two modules.

Firstly, the proposed model learns outfit compatibility with a recurrent neural network (RNN).
This is an architecture for processing sequential data that can be applied by considering outfits as sequences of fashion item features~\cite{han2017learning}. Following~\cite{han2017learning}, we use bidirectional long short-term memory (BiLSTM) as the RNN architecture. This module learns a sequence by evaluating the compatibility of local pairwise items.

Secondly, the proposed model extracts the outfit style with an autoencoder. 
By reducing item features included in an outfit, we obtained a style vector.
By encoding and decoding this vector using the autoencoder, we extracted a style representation common to many outfits.
Furthermore, we found that an outfit style can be represented by a linear combination of some typical style vectors.
Note that both sequence learning using Bi-LSTM and style extraction using the autoencoder are purely unsupervised methods. Moreover, these modules can be learnt end-to-end.

By incorporating a style extraction module in the previous work \cite{han2017learning}, we obtained the proposed model. We evaluated the effectiveness of this joint model for three tasks: a missing prediction task, a style extraction task, and an outfit generation task. 
The missing prediction task involves predicting the missing item in an input outfit. We observed that the proposed model outperformed the previous work. 
In the style extraction task, we found that the styles vectors extracted by the proposed model are interpretable. Besides, a style of an outfit can be represented by a linear combination of these styles vectors.
In the outfit generation task, the proposed model exhibited the ability to control the styles of generated outfits by adding style information. A summary of the outfit generation task is shown in Figure~\ref{fig:example}. Given an input item and a target style, the proposed model generates an outfit reflecting the style. In our experiments, we found that we could generate multiple outfits from the same item by changing the target style.

\section{RELATED WORK}
\subsection{Fashion Compatibility Learning and Outfit Generation}
Learning visual compatibility is an important problem in fashion recommendation. One possible solution is to learn the compatibility between a pair of fashion items using metric learning~\cite{mcauley2015image, veit2015learning, tautkute2018deepstyle, vasileva2018learning, shih2017compatibility}. To measure the compatibility between items, McAuley et al.~\cite{mcauley2015image} proposed a method of learning the relation between image features extracted by a pretrained convolutional neural network (CNN). Using a Siamese network, this feature extraction technique for compatibility learning was improved~\cite{veit2015learning, tautkute2018deepstyle}. These methods can learn complex relationships by merely providing samples of positive and negative pairs. However, they are not sufficient to represent the compatibility between fashion items. Their strategies map all fashion items to a common space, which does not appear to have sufficient flexibility to assess the distance between an arbitrary pair of items.

Vasileva et al.~\cite{vasileva2018learning} avoided these problems by learning similarity and compatibility simultaneously in different spaces for each pair of item categories. This method can make an outfit more fashionable by swapping each fashion item in the outfit for a better one. In contrast to these studies, Han et al.~\cite{han2017learning} adopted BiLSTM to learn the compatibility of an outfit as a whole. Their method successfully evaluated and generated outfits.

\subsection{Bidirectional LSTM}
LSTM is a variant of an RNN with memory cells and functional gates that govern information flow. It has been proven to be advantageous in real-world applications such as speech processing~\cite{graves2005framewise} and bioinformatics~\cite{hochreiter2007fast}. In fashion, LSTM can be effectively applied to understand Zalando's consumer behavior~\cite{lang2017understanding} and fashion recommendations~\cite{heinz2017lstm}. BiLSTM~\cite{graves2005framewise} combines the advantages of LSTM and bidirectional processing by scanning data in both the forward and backward directions.

\subsection{Fashion Style Extraction}
There have been some researches on extraction of fashion styles from co-purchase and outfit data~\cite{lee2017style2vec, liu2017deepstyle}. Lee et al.~\cite{lee2017style2vec} assumed that each item in an outfit and co-purchase item set share the same underlying fashion style. They proposed the  learning of the style by maximizing the probabilities of item co-occurrences. Liu et al. ~\cite{liu2017deepstyle} improved the recommendation performance on the basis of implicit feedback from users by decomposing a product image feature into a style feature and a category feature.

Takagi et al.~\cite{takagi2017makes} collected images corresponding to 14 modern fashion styles and trained a neural network to understand fashion style in a supervised manner. Although supervised methods are reliable, it is difficult to prepare accurate labels since an annotation task requires expert knowledge. Moreover, fashion items and snaps are not always labeled with clear words~\cite{simo2016fashion}. In contrast, Hsiao and Grauman \cite{hsiao2017creating} extracted underlying styles from unlabeled fashion images using unsupervised methods. 

Motivated by such researches, we aim to extract latent outfit styles using unsupervised methods. We adopt an aspect extraction method \cite{he-EtAl:2017:Long2}, which has succeeded in extracting aspects in the field of natural language processing. 

\section{METHODOLOGY}
\begin{figure*}
\includegraphics[width=\linewidth]{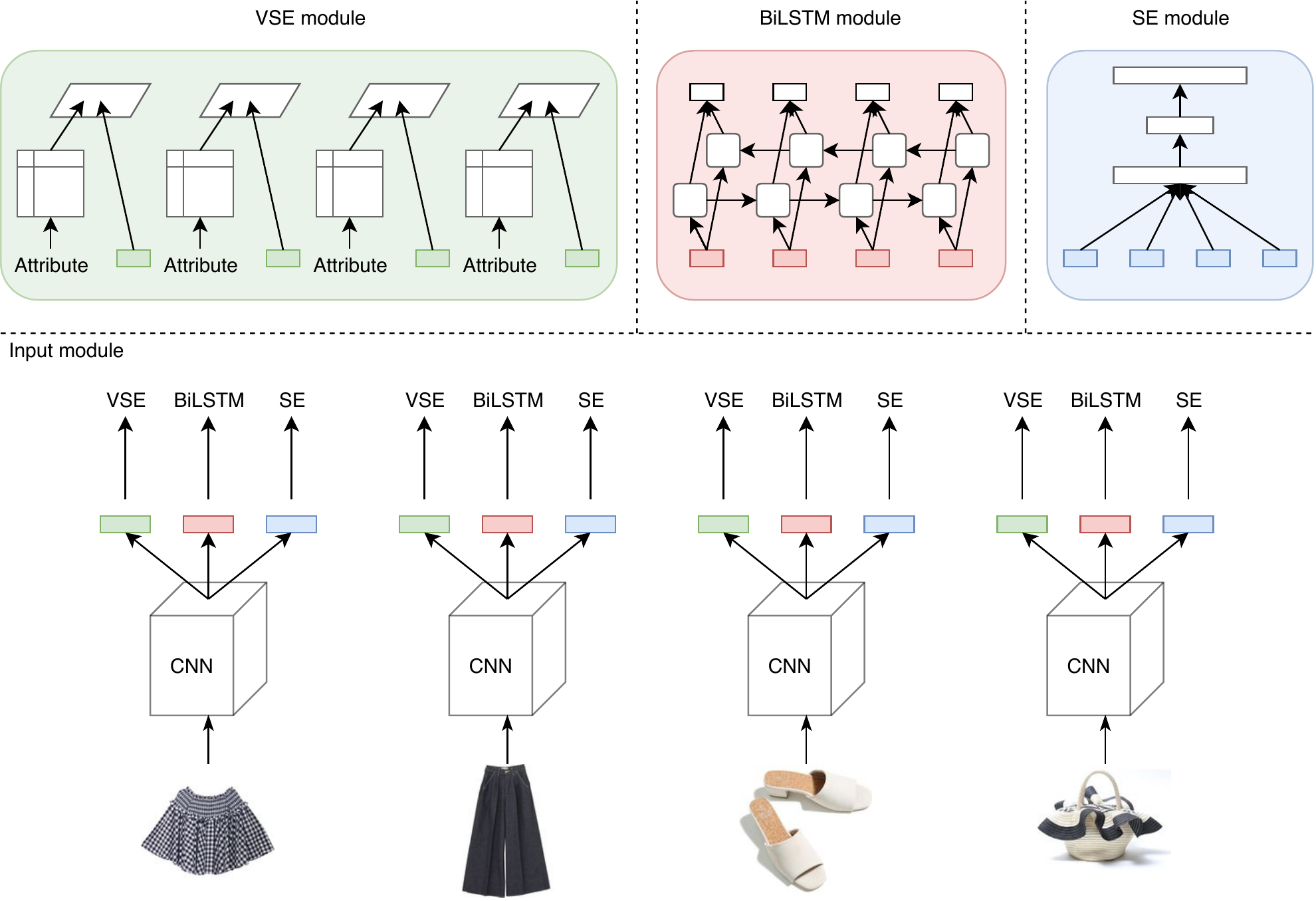}
\caption{Proposed architecture of the network. It consists of a CNN and three subnetworks. Item images in an outfit are fed through the CNN and encoded as image features. Then, the features are separately fed into the VSE, BiLSTM, and SE modules. The VSE module is only used for datasets in which items are labeled.}
\label{fig:proposed}
\end{figure*}

In this section, we describe the representation of outfit data, the modules used in the previous work and the proposed model, and the objective function of the proposed model. Figure~\ref{fig:proposed} illustrates the proposed architecture, where the visual-semantic embedding (VSE) module 
is only used for datasets in which items are labeled.

\subsection{Outfit Representation}
Following \cite{han2017learning}, we regard an outfit as a sequence of fashion items. A sequence $F=\{x_1,\ldots,x_{N}\}$ represents an outfit, where $x$ is a fashion item image feature extracted by a CNN and $N$ represents the length of a sequence. Note that $F$ has a variable length because the number of items in an outfit varies. The CNN is parameterized by $\theta_{CNN}$.

Fashion item categories such as tops and bottoms are associated with items constituting an outfit, and its priority with the order is determined according to the item category.

\subsection{Outfit Modeling with Bidirectional LSTM}
\label{bilstm}
BiLSTM is an extension of LSTM and is widely used to learn word sequences of natural language. In this model, an input sequence is processed simultaneously from both the forward and backward directions. By predicting the $t$th item in a sequence, BiLSTM can use all the information of items other than the $t$th item and thus it can perform more accurately than the commonly used LSTM \cite{han2017learning}.

\begin{eqnarray}
\lefteqn{ P(x_t |x_1, \ldots ,x_{t-1},x_{t+1}, \ldots ,x_N)  \nonumber }\\ 
&= P(x_t|x_1, \ldots ,x_{t-1}; \theta_{LSTMf})  P(x_t|x_N, \ldots ,x_{t+1}; \theta_{LSTMb})
\end{eqnarray}

A prediction using BiLSTM is solved by a joint model comprising a forward LSTM and a backward LSTM with parameters $\theta_{LSTMf}$ and $\theta_{LSTMb}$, respectively.

The forward LSTM maps a previous state $h_{t-1}$ and the next input $x_t$ to a current state $h_t$.

\begin{eqnarray}
i_t &=& \sigma(W_{xi}x_t + W_{hi}h_{t-1} + W_{ci}c_{t-1} + b_i) \nonumber \\
f_t &=& \sigma(W_{xf}x_t + W_{hf}h_{t-1} + W_{cf}c_{t-1} + b_f) \nonumber \\
c_t &=& f_t c_{t-1} + i_t \tanh(W_{xc}x_t + W_{hc}h_{t-1} + b_c) \nonumber \\
o_t &=& \sigma(W_{xo}x_t + W_{ho}h_{t-1} + W_{co}c_t + b_o) \nonumber \\
h_t &=& o_t \tanh(c_t) \nonumber
\end{eqnarray}
$W_{\alpha \beta}$ represents a weight matrix that maps vector $\alpha$ to vector $\beta$. $b$ represents a bias term. All weight matrices and biases are included in the forward LSTM parameters $\theta_{LSTMf}$. $\sigma$ represents the sigmoid function.

The state $h_t$ obtained by the above procedure is used to predict the next item by applying the softmax function:

\begin{eqnarray}
P(x_{t+1} | x_1, ...,x_{t}; \theta_{LSTMf}) = \frac{\exp(h_{t}x_{t+1})}{\sum_{x \in \chi} \exp(h_{t}x)}
\end{eqnarray}
where $\chi$ represents a set of candidate fashion items. Although $\chi$ originally contains all items in a dataset, this requires a huge computational cost. Thus, in the training phase, we built $\chi$ from items included in a mini batch.

The forward LSTM evaluates an outfit sequentially from the start to end. The loss for a given outfit $F$ can be calculated as:

\begin{eqnarray}
E_f(F; \theta_{LSTMf}) = -\frac{1}{N}\sum_{t=1}^{N} \log P(x_{t+1} | x_1, ..., x_{t}; \theta_{LSTMf})
\end{eqnarray}

The backward LSTM can be built in a similar way as :

\begin{equation}
P(x_t|x_{N},..., x_{t+1}; \theta_{LSTMb}) = \frac{\exp (\tilde{h}_{t+1} x_t)}{\sum_{x \in \chi}\exp (\tilde{h}_{t+1}x)} 
\end{equation}
\begin{equation}
E_b(F ; \theta_{LSTMb}) = -\frac{1}{N} \sum_{t=N -1}^{0} \log P (x_t | x_{N}, ..., x_{t+1};\theta_{LSTMb})
\end{equation}

\subsection{Visual-Semantic Embedding}
\label{vse}
VSE is a method for handling multimodal data. Simultaneously given a fashion item image and associated tags, VSE can provide a common expression between them.

A fashion item has a description that explains its characteristics and attributes. Let us denote it as $S=\{w_1,\ldots,w_M\}$, where $w_i$ represents a word included in the description. Given the one-hot vector $e_i$ corresponding to $w_i$, the mapping to an embedding space can be described as:

\begin{eqnarray}
v_i = W_T \cdot e_i \nonumber \\
v = \frac{1}{M} \sum_{i=1}^{M} v_i
\end{eqnarray}
where $W_T$ represents a word embedding matrix.
A mapping of the image feature to the embedding space is processed in the same way. Let $W_I$ denote the image-embedding matrix. Then,

\begin{eqnarray}
u = W_I \cdot x
\end{eqnarray}

It is preferable that $v$ and $u$ are mapped near to each other if they are derived from the same item. If they are from different items, it is preferable that they are mapped to separate positions. This relationship can be expressed using the hinge function:

\begin{align}
E_e(F, S; \theta_{VSE}) = & \sum_{u \in U} \sum_{v' \in V\backslash v^{(u)}} \max(0, m_s - d(u, v) + d(u, v')) \nonumber \\
& + \sum_{v \in V} \sum_{u' \in U\backslash u^{(v)}} \max(0, m_s - d(v, u) + d(v, u'))
\end{align}
where $d(u, v)$ represents the cosine distance. $U$ represents the set of embedded vectors $u$ and $U\backslash u^{(v)}$ represents the set of $u$ from which the $u$ corresponding to $v$ is removed. In other words, $U\backslash u^{(v)}$ are negative samples with respect to $v$. $V$ and $V\backslash v^{(u)}$ have the same definitions as $U$ and $U\backslash u^{(v)}$, respectively. $m_s$ is the margin of this hinge function. In the VSE module, $\theta_{VSE} = \{W_T, W_I\}$ are trainable parameters.

\subsection{Style Embedding with Autoencoder}
\label{style}
The previous work \cite{han2017learning}, in which the two modules described in \ref{bilstm} and \ref{vse} were joined, enabled us to learn item sequences and extract item features. However, this method may not be able to extract features of the whole outfit because it can only model the relationship between two neighboring items.

To address this problem, we introduce additional information, `style'.
In the fashion domain, style is one of the most important concepts in recognizing differences among outfits. It enables us to discriminate outfits and select clothes to wear. 
We aim to implement a model able to deal with style information and interpret outfit styles.

In natural language processing, document topics or semantics depend on the semantics of the words being used. Similarly, outfit styles depend on the styles of items in sequences.
To model this concept, we apply a module that has been proven successful in the field of natural language processing \cite{he-EtAl:2017:Long2} to fashion data.

Since outfit lengths vary, a feature extraction method from outfits is required to process a variable-length input. Li et al. \cite{Li2017MiningFO} proposed some methods to reduce a variable-length input. We adopt the mean as a reducing function:

\begin{eqnarray}
y_t =  W_s \cdot x_t \label{eq:style_reduce_1} \\
z = \frac{1}{N} \sum_{t}^{N} y_t \label{eq:style_reduce_2}
\end{eqnarray}
$y_t$ represents the style vector of an individual item and $z$ represents the style vector of an outfit. $W_s$ represents a weight matrix that maps an image feature to a style vector.

We obtained a representation for the style of an outfit. Next, through the process of compressing and reconstructing the style vector, we aim to obtain a basis of styles observed in a variety of outfits. Assuming that such a basis of styles exists, outfits can be represented as linear combinations of elements of the basis. For example, given a style basis that has two elements, `casual' and `formal', clothes or outfits can be labeled as casual, formal, or their mixture.

We use $p \in \mathbb{R}^{K}$ to denote the following mixture ratio:
\begin{eqnarray}
p = {\rm softmax}(W_z \cdot z + b_z) \label{eq:mixture}
\end{eqnarray}
where $W_z$ and $b_z$ are the weight matrix and bias vector used to map a style vector to a mixture ratio. Since $p$ is assumed to be a mixture ratio, the softmax function is applied so that each element of $p$ is nonnegative and the sum of the elements is 1. $K$ represents the number of elements of the style basis.

Next, we consider the reconstruction of style vector $z$. We denote this process as a linear combination of style embeddings and the mixture ratio,

\begin{eqnarray}
r = W_p^T \cdot p \label{eq:combine}
\end{eqnarray}
where $W_p$ is a style-embedding matrix. Each row of this matrix corresponds to an element of the basis. 

Since $r$ is the reconstructed style vector, we design the following objective function to match $r$ and $z$, which is similar to an autoencoder:

\begin{eqnarray}
E_s(F; \theta_{SE}) = \sum_{r \in R} \sum_{z' \in Z\backslash z^{(r)}} \max(0, m_r - d(r, z) + d(r, z'))
\end{eqnarray}
where the definitions of $d(r, z)$, $r \in R$, $z' \in Z\backslash z^{(r)}$, and $m_r$ are similar to those in \ref{vse}.

Assuming that each row of $W_p$ is an element of the basis, the regularization applied for each row to become unique is

\begin{eqnarray}
E_r(W_p) = || W_{pn} W_{pn}^T - I||
\end{eqnarray}
where $W_{pn}$ represents the normalized $W_p$ over the rows and $I$ is the identity matrix. $\theta_{SE} = \{W_s, W_z, b_z, W_p\}$ are trainable parameters.

\subsection{Training Objective}
The proposed model consists of three modules, the BiLSTM (\ref{bilstm}), VSE (\ref{vse}), and style embedding (SE) modules (\ref{style}). The overall objective function is obtained as follows by adding the objective function of each module:

\begin{eqnarray}
\min_{\Theta} \sum_{F}(E_f + E_b) + \lambda_e E_e + \lambda_s E_s + \lambda_r E_r
\end{eqnarray}
where $\Theta = \{\theta_{LSTMf}, \theta_{LSTMb}, \theta_{VSE}, \theta_{SE}\}$. $\lambda_{e}$, $\lambda_{s}$, and $\lambda_r$ are hyperparameters that control the weight of each objective function.

One of the benefits of the proposed model is that it can be applied to data that is  lacking attributes. Because the SE module can be trained in an unsupervised manner, it requires no attributes. When learning the model with only an outfit sequence $F$, its objective function consists of BiLSTM and SE.

\begin{eqnarray}
\min_{\Theta} \sum_{F}(E_f + E_b) + \lambda_s E_s + \lambda_r E_r
\end{eqnarray}

Although the model trained without any attributes cannot manipulate outputs expressly using attributes, it enables the styles of outputs to be controlled with the mixture ratio $p$. This ability is shown in \ref{outfit_generation}.

\section{EXPERIMENTAL RESULTS}

In this section, we describe the experimental settings and report the results. We evaluate the proposed model for three tasks: a missing prediction task, a style extraction task, and an outfit generation task.

\subsection{Datasets}
\subsubsection{Polyvore}
Polyvore (www.polyvore.com) is a now defunct popular fashion website where users could create outfits and upload them. To evaluate the effectiveness of our methods compared with previous work, we used the Polyvore dataset in our testing. It includes 164,837 items of clothing grouped in 21,889 outfits. Each item is labeled multiple tags indicating its attributes. For details, refer to \cite{han2017learning}.

\subsubsection{IQON}
IQON (www.iqon.jp) is a fashion web service for women. Like Polyvore, IQON users can create original outfits by collaging images of fashion items. We have created an IQON dataset for outfit generation tasks. The IQON dataset consists of recently created and high-quality outfits that acquired 50 or more votes from other users and were created from January 1, 2016, to January 1, 2018. It includes 199,792 items grouped in 88,674 outfits. These outfits are split into 70,997 for training, 8,842 for validation, and 8,835 for testing. Unlike the Polyvore dataset, none of the items in the dataset are labeled with tags.

The dataset is similar to the Polyvore dataset: however, the styles of the included outfits are different because the main users are Japanese women. To test the generalization performance of our method, we evaluated models with multiple datasets.

\subsection{Implementation}
We adopted Inception-V3 \cite{DBLP:journals/corr/SzegedyVISW15} to map a fashion image to a 2048-dimensional (2048D) feature vector. The dimensions of the other variables and parameters are shown in Table \ref{tab:params}. The hyperparameters of the objective function $\lambda_e$ ,$\lambda_s$, and $\lambda_r$ are equal to $1.0$, $1.0$, and $0.1$, respectively. The other settings follow \cite{han2017learning}.

\begin{table}
  \caption{Parameter details}
  \label{tab:params}
  \begin{tabular}{ccl}
    \toprule
    Module&Parameter and Variable&Dimension\\
    \midrule
    BiLSTM & $x_t$ & $\mathbb{R}^{512}$ \\
     & $h_t$ & $\mathbb{R}^{512}$ \\
    VSE\footnotemark[1] & $W_T$ & $\mathbb{R}^{2757 \times 512}$ \\
     & $W_I$ & $\mathbb{R}^{2048 \times 512}$ \\
    SE & $W_s$ & $\mathbb{R}^{2048 \times 256}$\\
     & $W_z$ & $\mathbb{R}^{256 \times 8}$ \\
     & $b_z$ & $\mathbb{R}^{8}$ \\
     & $W_p$ & $\mathbb{R}^{8 \times 256}$ \\
  \bottomrule
\end{tabular}
\end{table}
\footnotetext[1]{This module was only used for the experiment using the Polyvore dataset.}

\subsection{Missing Prediction}
Fill in the blank (FITB) is a task predicting an item removed from an outfit created by a human. Given an outfit lacking an item, the evaluated models choose the most compatible item from the candidate items sampled at random. This task evaluates a compatibility recognition performance between an item sequence and an item alone. In this experiment, the number of candidate items is set to four, and then negative items are sampled from outfits excluding the query outfit.

The previous work evaluated the compatibility as follows:

\begin{align}
\label{eq:fitb}
x_a = {\rm argmax}_{x_c \in C} \frac{\exp (h_{t-1} x_c)}{\sum_{x \in C} \exp (h_{t-1}x)} + \frac{\exp (\tilde{h}_{t+1} x_c)}{\sum_{x \in C} \exp (\tilde{h}_{t+1}x)}
\end{align}
where the first and second terms on the right side are the scores calculated by the forward LSTM and backward LSTM, respectively. $C$ represents the set of candidate items.

Although the previous work computed scores with BiLSTM, the proposed model adds another measure to the previous work. Assuming that a user-created outfit has a unified style, it is considered that the style of each item included in the outfit is similar.  To compare styles, we denoted the style similarity as follows:

\begin{eqnarray}
SS(F_{s1},F_{s2}) = d(z_{s1}, z_{s2}) \label{eq:ss}
\end{eqnarray}

We used Eqs. \ref{eq:style_reduce_1} and \ref{eq:style_reduce_2} to map $F$ to $z$.

On this basis, we redefine the compatibility evaluator as,

\begin{align}
x_a =\  & {\rm argmax}_{x_c \in C} \frac{\exp (h_{t-1} x_c)}{\sum_{x \in C} \exp (h_{t-1}x)} \nonumber \\
& + \frac{\exp (\tilde{h}_{t+1} x_c)}{\sum_{x \in C} \exp (\tilde{h}_{t+1}x)} 
+ \gamma SS(F, \{x_c\})
\end{align}
where $F$ represents an input sequence with an item removed and $\gamma$ is a hyperparameter that represents a weight of style similarity.

\begin{table}
  \caption{Performance comparison among different datasets and models in terms of FITB accuracy.}
  \label{tab:fitb}
  \begin{tabular}{cccl}
    \toprule
    Dataset&Method&$\gamma$&Acc\\
    \midrule
    Polyvore & Bi-LSTM + VSE \cite{han2017learning} & - & 0.726 \\
    &Bi-LSTM + SE (this paper) & 0.0 & 0.729 \\
& & 0.2 & 0.727 \\
& & 0.5 & 0.723 \\
&Bi-LSTM + VSE + SE (this paper) & 0.0 & 0.728 \\
& & 0.2 & \textbf{0.732} \\
& & 0.5 & \textbf{0.732} \\
\hline
IQON&Bi-LSTM & - & 0.703 \\
&Bi-LSTM + SE (this paper) & 0.0 & \textbf{0.715} \\
& & 0.2 & 0.713 \\
& & 0.5 & 0.711 \\
  \bottomrule
\end{tabular}
\end{table}

The results are shown in Table \ref{tab:fitb}. The proposed models outperformed previous models in the experiment for both the Polyvore and IQON datasets. In particular, we observed that the accuracy of BiLSTM + SE is nearly equal to that of BiLSTM + VSE. This suggests that the proposed model can achieve the same accuracy without the attribute information $S$.
We also observed that the BiLSTM + SE model without the style similarity term ($\gamma = 0$) is superior to the model with the style similarity term ($\gamma > 0$). In contrast, the BiLSTM + VSE + SE model showed the opposite result. These results indicate that the simultaneous use of BiLSTM and SE improves the FITB accuracy.

\subsection{Style Extraction}
\label{outfit_clustering}
\begin{figure*}
\includegraphics[width=\linewidth]{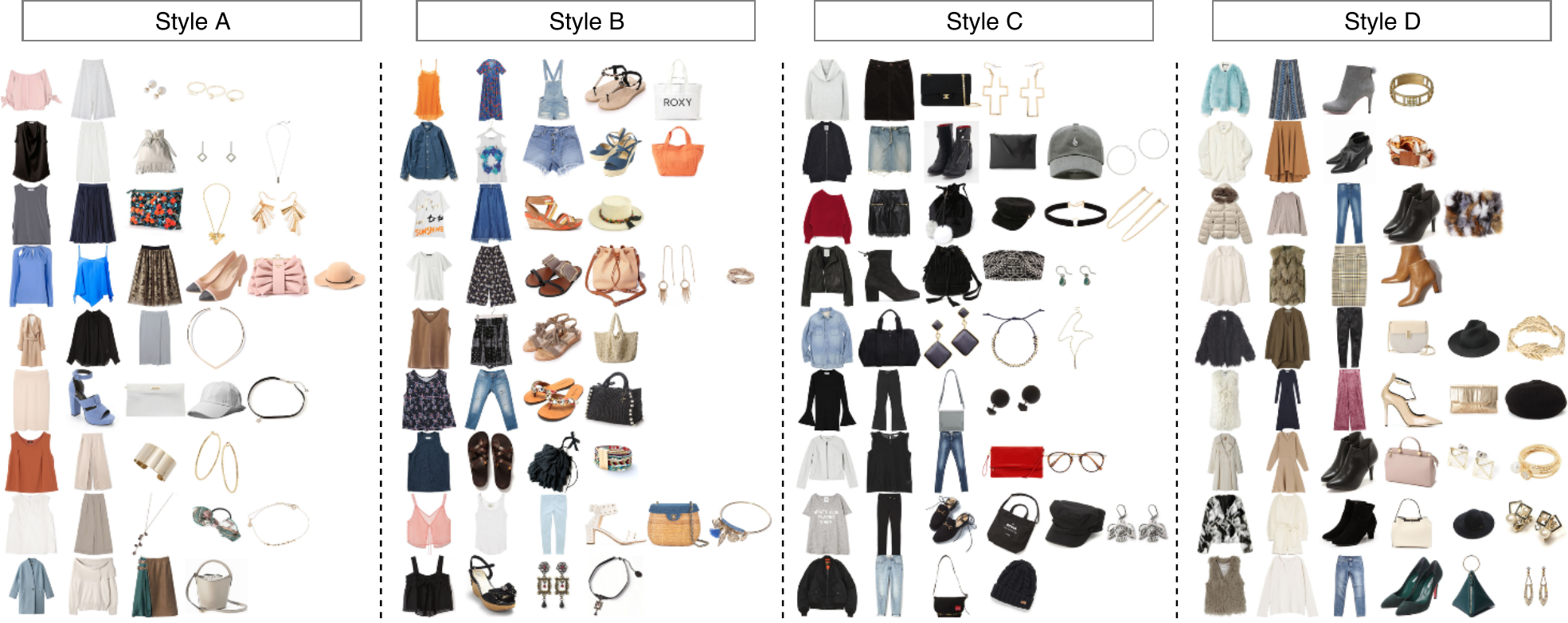}
\caption{Examples of some elements of a style basis.}

\label{fig:outfit_clustering}
\end{figure*}

In \ref{style}, we assumed that an outfit can be represented as a mixture of elements of a style basis. The mixture ratio $p$ is a weight for the corresponding basis. Similarly, an element of the style basis itself can be encoded as a one-hot vector in the space in which the mixture ratio of each outfit is embedded. To investigate a visual characteristic of the elements of a style basis, we visualized some typical outfits near each element of a style basis in Figure \ref{fig:outfit_clustering}. We used the IQON dataset and BiLSTM + SE model in this experiment.

\begin{figure}
\includegraphics[width=8cm]{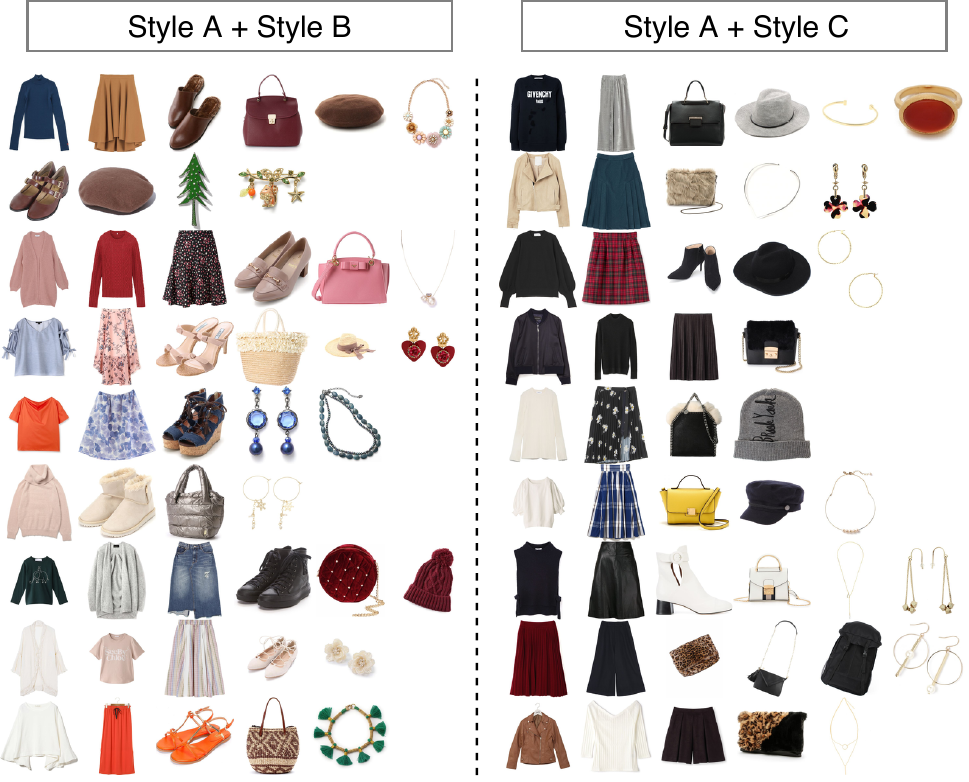}
\caption{Examples of some styles combined two elements of style basis.}

\label{fig:outfit_clustering2}
\end{figure}

In Figure \ref{fig:outfit_clustering}, each row corresponds to an outfit created by a user. The four styles, that are the part of the style basis extracted by the proposed model, have the following features.

\begin{itemize}
\item {\bf Style A} These outfits have a gentle color scheme based on pale tones reminiscent of spring or early summer. The main items of this style are skirts or simply designed items.
\item {\bf Style B} This style is dominated by sandals and jeans, which may be associated with summer. It has high-contrast colors compared with Style A.
\item {\bf Style C} This style contains a lot of dark and black items. Since there are many items made with leather or metal, this style suggests strength and individuality.
\item {\bf Style D} This style consists of winter outfits. Some outer clothes are made with thick fabrics, such as fur. In addition, there are high-heel boots and gold accessories.
\end{itemize}

These visualized styles can be easily distinguished without expert knowledge of fashion. This suggests that the style basis has a certain ability to represent outfit styles that can be interpreted.

In Figure \ref{fig:outfit_clustering2} shows two examples of mixture that are calculated by combining two elements of the style basis. We used Eq. ~\ref{eq:combine} to obtain these mixtures. `Style A + Style B' is equal to $0.5 \cdot \text{Style A} + 0.5 \cdot \text{Style B}$. This style contains skirts that are also observed in Style A, whereas the contrast of the item colors is high like Style B. `Style A + Style C' is equal to $0.6 \cdot \text{Style A} + 0.4 \cdot \text{Style C}$. This style also contains a lot of skirts like Style A, whereas its color tone is similar to Style C. These results indicate that the mixture representation of style vectors makes a complex style of an outfit a simple expression.

\subsection{Outfit Generation}
\label{outfit_generation}
We demonstrate our approach to generating outfits. In the generation task using the proposed model, the BiLSTM module generates a sequence from query item images and the SE module adjusts the global style of the generated sequence.
To simultaneously evaluate the likelihoods of the sequence and global style, we defined the score function as follows:

\begin{equation}
\begin{split}
score(F|p_{target})
&= -(E_f(F; \theta_{LSTMf}) + E_b(F; \theta_{LSTMb})) \\
&\ \ \ + \beta d(W_p^T \cdot p_F, W_p^T \cdot p_{target}) \label{eq:seqscore}
\end{split}
\end{equation}
where $p_{target}$ represents the target style, which is a parameter of the function. $p_F$ represents the style of the query sequence computed using Eqs. \ref{eq:style_reduce_1}-\ref{eq:mixture}. $d(W_p^T \cdot p_F, W_p^T \cdot p_{target})$ represents the similarity between the style of the generated sequence and the target style. This term makes it difficult to select items that do not suit the given style. $\beta$ is a hyperparameter used to control the balance between the sequence likelihood and the style similarity.

We adopted a beam search to maximize the total score \cite{Vinyals2015ShowAT}. First, given a query image or image sequence, the BiLSTM module predicts some previous and following items by computing Eq. ~\ref{eq:fitb}. By combining the query item (sequence) with the predicted items, we can obtain candidate sequences. Next, given a target style $p_{target}$, the BiLSTM module and SE module compute the score of each sequence by computing Eq.~\ref{eq:seqscore}.

\begin{figure*}
\includegraphics[width=16cm]{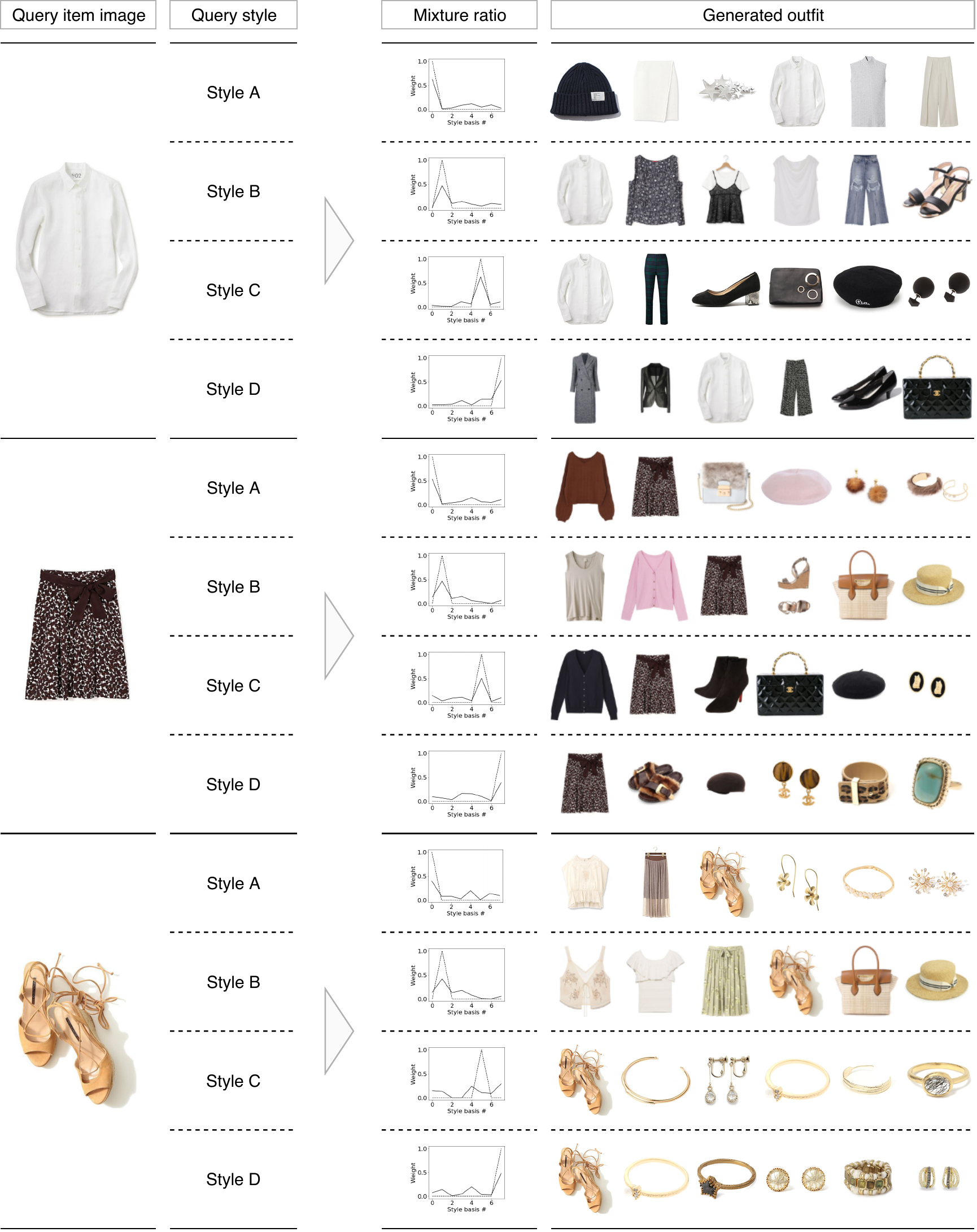}
\caption{Example of style-controlled outfit generation. The input is a query item and a target style. The output is the corresponding outfit including the query item. The sparkline is a comparison between the target style and the style of the generated outfit.}
\label{fig:outfit_generation}
\end{figure*}

Figure \ref{fig:outfit_generation} shows three examples and results of the outfit generation task. The first column is the input image. Our model attempts to generate a set of items compatible with the input item. The next column is the target styles. It is desirable that the generated outfit reflects these styles. Styles A, B, C, and D are the same as those in Figure~\ref{fig:outfit_clustering}. The sparklines in the next column show the mixture ratios. The solid line is the mixture ratio of the generated outfit $p_F$ and the dashed line is the target mixture ratio $p_{taraget}$ in Eq. \ref{eq:seqscore}. The final column is the generated outfit.

In the first example with the white shirt, the generated outfits generally reflect each input style. In particular, the outfit reflecting Style B contains jeans and sandals, and the one reflecting Style D contains a thick coat.
These results are consistent with the characteristics of each style visualized in \ref{outfit_clustering}.
 In the second example with the skirt, it can also be confirmed that the generated outfits reflect the characteristics of the style. For example, the outfit reflecting Style B contains items reminiscent of summer such as sandals and a straw hat. The outfit reflecting Style C predominantly contains black items.
 The third example is an example of a failure. When targeting Styles C and D, our model fails to generate an outfit. This is because the style of the input image does not suit the target style. These results imply that a large difference between the style of the input image and the target style makes it difficult to generate a natural outfit.

\section{CONCLUSION}
We have introduced the concept of `style' into an outfit recognition task and proposed a model able to learn the outfit sequence and style simultaneously in an unsupervised manner.
We applied the proposed model to outfit datasets to evaluate its efficiency. 
In a prediction task, the proposed model outperformed the previous model. In a generation task, our model succeeded in generating outfits with a typical style.
These results indicate that the proposed model can assess fashion item compatibility and outfit style.

A future work is to develop a recommender system. 
Given a personal collection of fashion items, such as those on a Pinterest Board or Amazon Wish List, the proposed model can extract their styles, which can be regarded as personal style preferences. Thus, the proposed model will be able to recommend items and outfits that reflect the preferred styles of individual users.

\bibliographystyle{ACM-Reference-Format}
\bibliography{sample-bibliography}

\end{document}